# SELF-LEARNING AI FRAMEWORK FOR SKIN LESION IMAGE SEGMENTATION AND CLASSIFICATION

Anandhanarayanan Kamalakannan[1], Shiva Shankar Ganesan[2] and Govindaraj Rajamanickam[1]

[1]Department of Cyber Physical Systems, Central Electronics Engineering Research Institute, CSIR Madras Complex, Taramani, Chennai, India
[2]PG Scholar, Dept. of Computer Science, University of Bridgeport, 126 Park Ave, Bridgeport, USA

*ABSTRACT*

*Image segmentation and classification are the two main fundamental steps in pattern recognition. To perform medical image segmentation or classification with deep learning models, it requires training on large image dataset with annotation. The dermoscopy images (ISIC archive) considered for this work does not have ground truth information for lesion segmentation. Performing manual labelling on this dataset is time-consuming. To overcome this issue, self-learning annotation scheme was proposed in the two-stage deep learning algorithm. The two-stage deep learning algorithm consists of U-Net segmentation model with the annotation scheme and CNN classifier model. The annotation scheme uses a K-means clustering algorithm along with merging conditions to achieve initial labelling information for training the U-Net model. The classifier models namely ResNet-50 and LeNet-5 were trained and tested on the image dataset without segmentation for comparison and with the U-Net segmentation for implementing the proposed self-learning Artificial Intelligence (AI) framework. The classification results of the proposed AI framework achieved training accuracy of 93.8% and testing accuracy of 82.42% when compared with the two classifier models directly trained on the input images.*

## 1. INTRODUCTION

Deep learning techniques are widely used in medical imaging application for implementing clinical-decision support algorithms. Here, we establish a decision support algorithm based on two-stage deep learning framework for screening of skin lesion images. To perform effective feature extraction and classification on the lesion images, background skin portion needs to be discarded before training the algorithm. Thus, image segmentation technique plays a vital role in extracting Region of Interest (RoI) portion from an image. For segmentation, deep learning algorithms generally require big image dataset with manual annotation [1]. The medical image dataset mostly do not have annotation information for the entire voluminous data. To resolve this issue and train U-Net segmentation model, author [2] has discussed the implementation of ground truth mask generation using fundamental image processing method. This method consists of adaptive histogram equalization and morphological processing techniques for generating initial binary labels for the segmentation. The approximate annotation information for an unlabelled image dataset can be obtained using combination of different processing methods like segmentation, clustering, morphological operations and selective search etc. For skin lesion image segmentation, K-means clustering algorithm is found to be an appropriate method with fixed number of clusters [3, 4]. In this framework, a self-learning annotation scheme was proposed for generating initial annotation information using unsupervised K-means clustering algorithm. The generated labels were used to train the U-Net segmentation model. The trained model segments the lesion region and sends the cropped portion to Convolutional Neural Network (CNN) classifier. The classifier model further classifies the segmented portion as benign or malignant.

In recent years, many methods were discussed for automatic skin lesion segmentation and classification for dermoscopic images. S. M. Jaisakthi et. al [5] proposed an automatic segmentation algorithm for skin lesion images using the combination of GrabCut and K-means clustering, in which GrabCut segments the foreground image and K-means along with flood-fill technique extracts the lesion region from the foreground. Youssef Filali et. al [6] trained SVM algorithm to classify the skin lesion images with structural and textural features extracted from the segmented portion. Manu Goyal .et al [7] proposed an ensemble deep learning method for lesion image segmentation, in which the results of two models namely DeepLabv3 and Mask-RCNN were combined to produce accurate segmentation mask. Similarly for classification, the deep learning models are rigorously trained on medical image dataset like chest X-rays to classify images infected with signs of bacterial and viral infections. These models are also trained on retinal optical coherence tomography (OCT) images to classify age-related macular degeneration and diabetic retinopathy [8]. The trained CNN model is also effective in distinguishing normal OCT images from age-related macular degeneration [9].

In this work, we propose a self-learning AI framework consisting of two-stage deep learning model for performing skin lesion image segmentation and classification. The first stage discusses about how to segment region of interest in lesion images without ground truth information. The second stage explains about the benefits of image segmentation in improving the classification accuracy of LeNet-5 classifier model. The results of the proposed framework in classifying benign and malignant lesion images are tabulated and compared with the CNN models namely ResNet-50 and LeNet-5 trained directly on input images. The rest of the paper is organized as follows: Section 2 briefly explains about the proposed two-stage deep learning framework for skin lesion image segmentation and classification. Section 3 describes about the image dataset preparation and algorithm implementation. The results of the proposed algorithm with other CNN classifier models are reported and discussed in Section 4. The conclusion on the self-learning AI framework and its implementation is discussed in Section 5.

## 2. SELF-LEARNING AI FRAMEWORK

The annotation information is very much essential for training any deep learning segmentation algorithm. The main objective is to develop a two-stage deep-learning AI framework which can segment lesion region from its normal portion and classifying them under two main classes namely benign and malignant. The segmentation model in the AI framework learns the annotation information in a self-learning manner. The performance of the framework is evaluated and compared with single stage CNN classifier models.

### 2.1. Self-learning Annotation Scheme

Self-learning annotation scheme is based on unsupervised segmentation approach. In this method, to generate initial annotation data for skin lesion image segmentation, spatial K-means clustering algorithm was used. The K-means algorithm classifies the image pixels by grouping similar pixel intensities into i number of clusters. Based on the need, the optimum i-value was chosen. In skin images, for lesion region segmentation, we set i=5 to obtain five cluster regions covering entire image [4]. We need only two clusters representing the ROI and the background [3]. Therefore the five clusters are grouped under two regions to obtain binary segmentation mask as depicted in the flowchart Fig.1. The generated binary mask for each lesion image is manually checked for their approximate coverage over the lesion region. In case, the background region is covered by the binary mask instead of the lesion region, then it is simply inverted using open-source image editor like GIMP (GNU Image Manipulation Program). The final set of approximately corrected binary segmentation mask becomes initial annotation data and serves as target labels for training U-Net segmentation model.

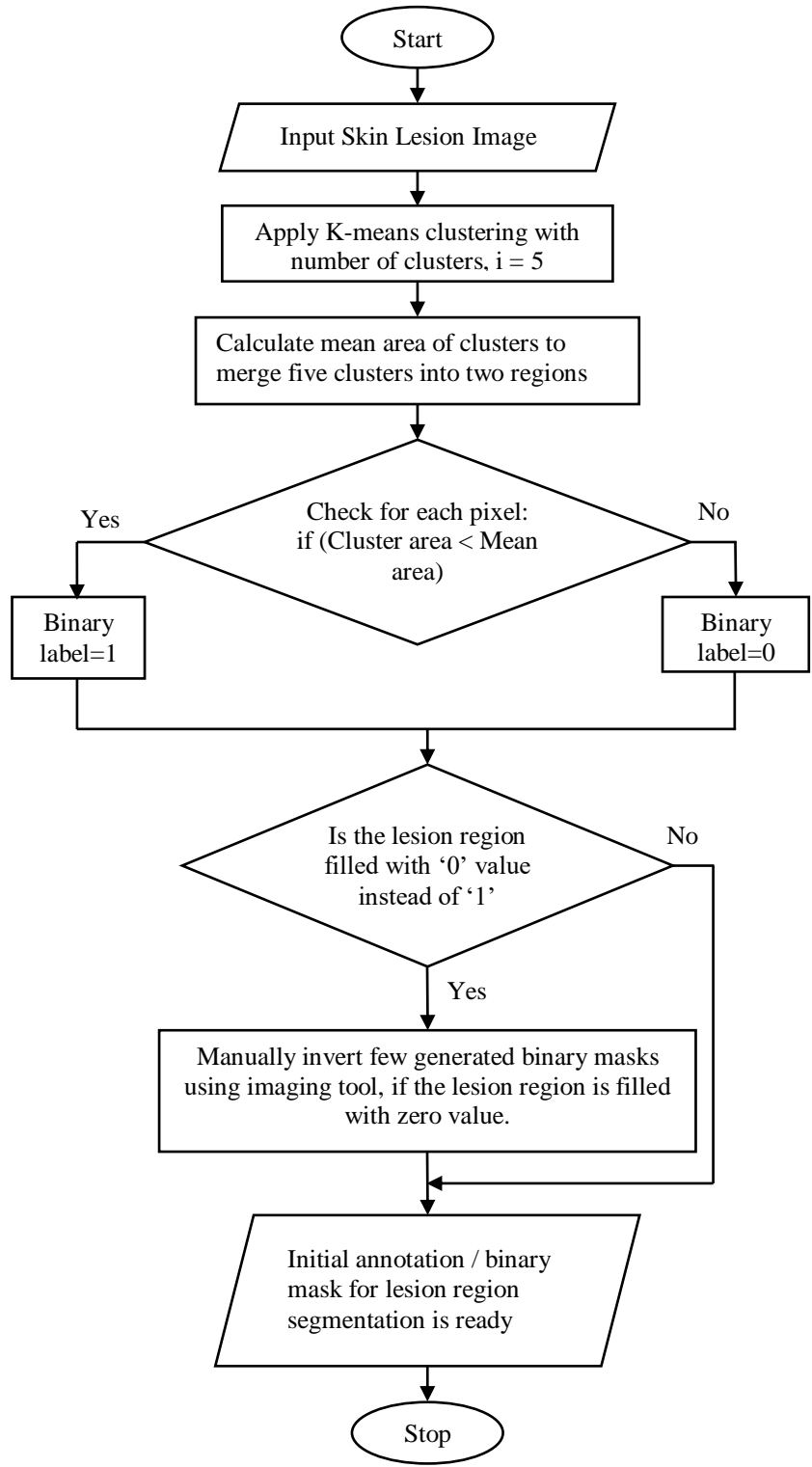

Figure 1. Flowchart of the self-learning annotation scheme

The self-learning annotation scheme has reduced the time and expertise needed for manual labeling. This scheme exploits the outcome of unsupervised methods to effectively train deep learning based segmentation techniques [1] and also helps to deal with medical image dataset which don not have ground truth information.

## 2.2. U-Net Segmentation

After generating the initial annotation data using self-learning annotation scheme from the lesion dataset, the U-Net model was used as the deep learning architecture to implement lesion region segmentation. In medical image segmentation, the U-Net model generally poses better performance in terms accuracy and efficiency [3, 10]. The U-Net architecture consists of three main sub-divisions: Down-sampling layer, Up-sampling layer and skip connection. The down-sampling layer consists of four convolution layers for encoding the image data for analysis and up-sampling layer consists of four convolution layers for decoding the latent space for image reconstruction. Skip connections are used to transfer fine-grained information from the low-level layers of the encoding path to the high-level layers of the decoding path to enhance the contextual information for image segmentation [1, 2].

To train U-Net segmentation model, the skin lesion images were resized to dimension of 256 x 256 pixels and given as input to the segmentation model along with target binary mask generated by annotation scheme. The training mainly focusses on distinguishing lesion region from normal skin portion for image segmentation. Image segmentation is a binary class problem, where the lesion region needs to be separated from the normal skin portion and the background. Thus the U-Net training model was optimized using adam optimizer with binary cross-entropy as loss function. After training several epochs, the U-Net model showed better segmentation performance than the K-means clustering algorithm as shown in Fig.2. The developed model was used to segment the lesion region from the skin image.

## 2.3. Lesion Image Classification

Different CNN architectures like LeNet-5, VGG-11, VGG-16 & ResNet-50 etc. were used for image classification. Based on the application requirement these architectures can be trained with different image datasets. A. Romero Lapez [11] explains about the implementation of a two-class VGG-16 model to classify skin lesion images in to two main categories viz. benign and malignant using transfer learning paradigm. The trained VGG model achieved 81.33% test accuracy. In the similar way, to classify the skin lesion images, the ResNet-50 and LeNet-5 CNN architectures were chosen as classifiers and their accuracies were analysed with training dataset before and after lesion region segmentation.

## 2.4. Proposed Skin Lesion Image Analysis algorithm

To provide complete solution from unsupervised segmentation to classification for the lesion dataset, a two-stage deep learning framework was proposed [12]. First stage is the U-Net segmentation model along with self-learning annotation scheme and second stage is the classification model as shown in the block diagram Fig.3. The segmentation algorithm provides best solution for lesion region segmentation on the image dataset which do not have any ground truth information. The self-learning annotation scheme generates initial labeling information for training the U-Net model. The proposed algorithm takes skin lesion colour image as input and passes it to trained U-Net model to obtain segmented output as binary mask. The binary mask performs AND operation with the skin image to crop the lesion region. The cropped lesion region images were used to train the LeNet-5 classifier model for further classification. During testing, the cropped region is passed to trained CNN model for classification. Finally the segmented lesions are either classified as benign or malignant.

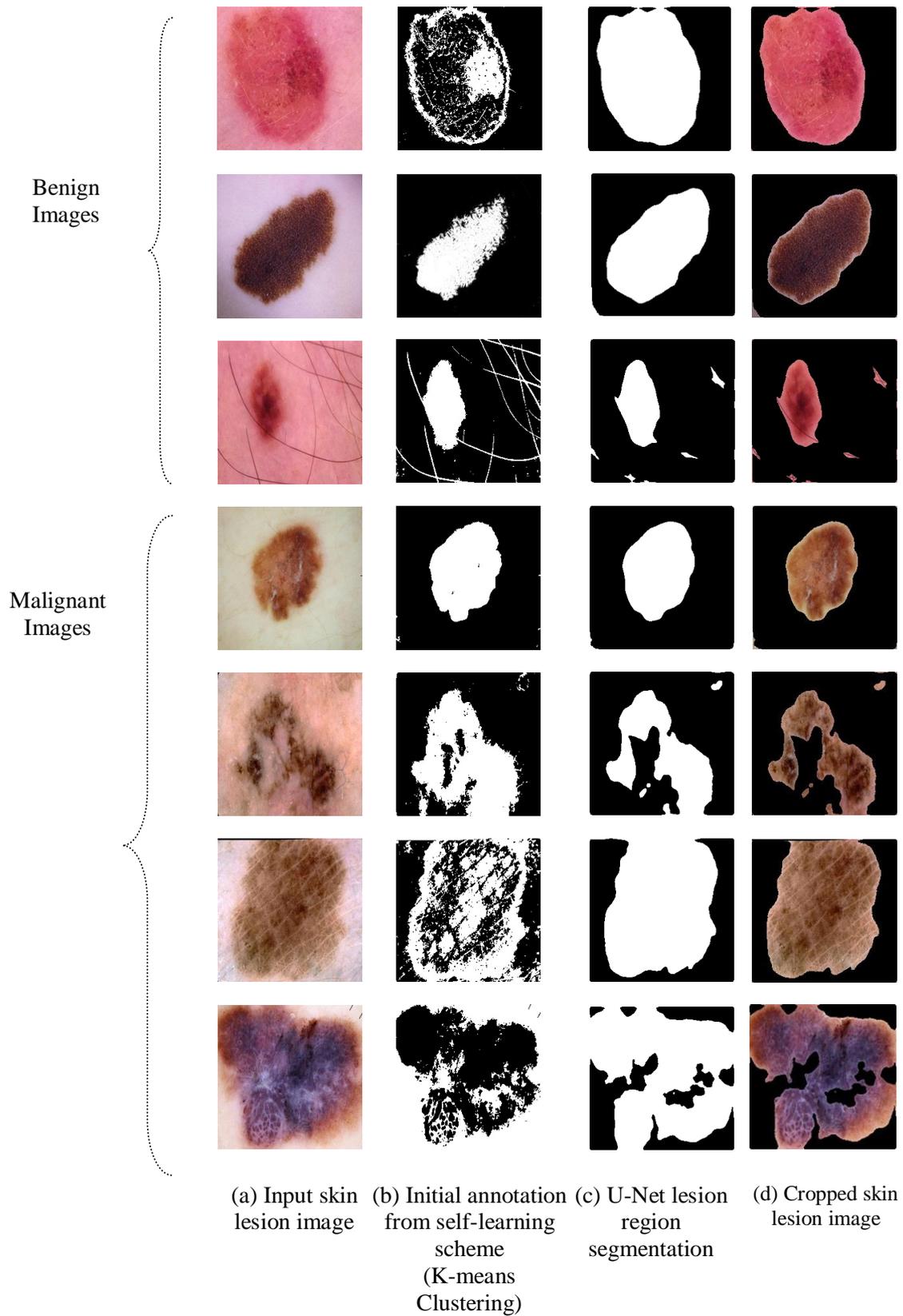

(a) Input skin lesion image  (b) Initial annotation from self-learning scheme (K-means Clustering)  (c) U-Net lesion region segmentation  (d) Cropped skin lesion image

Figure 2  Lesion region segmentation results of U-Net model trained with initial annotation obtained from self-learning scheme

The two stage deep learning algorithm enables the classification model to extract more representative and specific features [12] from segmented region instead of the whole image, which improves the overall classification accuracy of the model as discussed in Section 4.

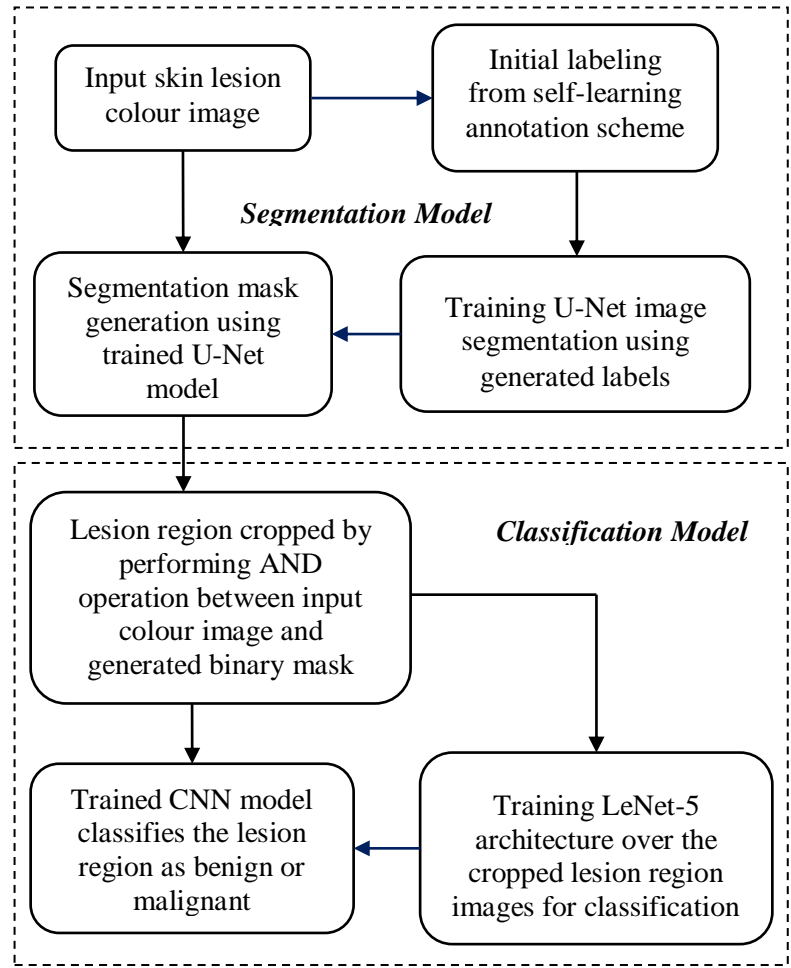

Figure 3 Block diagram of proposed self-learning AI framework for skin lesion region segmentation and classification

## 3. ALGORITHM IMPLEMENTATION

### 3.1. Image Dataset Preparation

International Skin Imaging Collaboration (ISIC) has developed an image archive [13, 14], which is an international repository of dermoscopic images consisting of different diagnostic category images namely melanoma, melanocytic nevus, basal cell carcinoma, actinic keratosis, benign keratosis (solar lentigo / seborrheic keratosis / lichen planus-like keratosis), dermatofibroma, squamous cell carcinoma and vascular lesion.

Based on the category, all these images can be grouped in to two main classes namely benign (non-cancerous) and malignant (cancerous). Melanocytic nevus, benign keratosis, dermatofibroma and vascular lesion category images belong to benign class whereas melanoma, basal cell carcinoma, actinic keratosis and squamous cell carcinoma lesion category images belong to malignant class. The training and testing images were chosen from the ISIC dataset which has more than 10000 lesion images. Based on the image category, 1460 lesion images were grouped under benign class and 1400 lesion images were identified under malignant class.

## 3.2. Implementation Details

Both the U-Net segmentation and CNN classifier models were implemented using Keras library with TensorFlow as backend. U-Net segmentation model was trained from scratch using the skin lesion images chosen from the ISIC dataset. To obtain initial labels for U-Net training, self-learning annotation scheme was coded in python language with OpenCV library support. The self-learning annotation scheme uses K-means clustering algorithm with grouping conditions to produce initial binary segmentation mask for the chosen image dataset. Around 2500 lesion images comprising both classes along with binary masks were used to train the U-Net model. The U-Net was trained for 25000 iterations to obtain optimum image segmentation result.

Two different CNN classifier models namely ResNet-50 and LeNet-5 were trained and tested with the lesion image dataset before and after image segmentation using the U-Net model. To train these classifiers, 2200 images distributed equally among two classes were taken. After training, the classifiers were tested with the remaining 360 benign and 300 malignant lesion images. To implement the proposed algorithm, the entire training and testing image dataset was passed to trained U-Net segmentation model and obtained the segmented lesion images for training and testing the LeNet-5 classifier model.

Both the ResNet-50 and LeNet-5 classifier models were trained and tested on the lesion image dataset for comparison. All the classifier models were trained for optimum number of epochs. The trained model weights of U-Net and CNN classifiers were saved in HDF5 format. Finally the proposed model segments the lesion region and classifies the region as benign or malignant. The training and testing accuracy of two CNN classifier models and the proposed algorithm were evaluated based on the performance metrics and reported in Section 4. The developed deep learning algorithms were trained and tested on a workstation PC with Intel Xeon E5-2620 v3 2.40 GHz processor, 16 GB RAM, and Nvidia Quadro P4000 graphics card with 8GB memory.

## 4. RESULTS & DISCUSSION

To propose self-learning AI framework for skin lesion image dataset, the U-Net segmentation and LeNet-5 CNN models were used to segment lesion regions and classify them as benign or malignant. To perform testing of different classifiers on skin lesion images, a total of 660 images comprising 360 benign and 300 malignant were considered as test dataset. The classifier results of LeNet-5, ResNet-50 and the proposed algorithm on the test dataset are presented in the form of confusion matrix as shown in Table-1. After testing the two classifier models, it was observed that the training and testing accuracy of LeNet-5 was better than the ResNet-50 as shown in Table-2. Based on the performance, LeNet-5 was considered as classifier model for the proposed self-learning algorithm.

For lesion region segmentation, already a proven U-Net medical image segmentation model was used [1, 2, 3 & 10]. To train the U-Net model, self-learning annotation scheme was deployed to initiate generation of labeling information for the image dataset which don't have ground truth. The segmentation results thus obtained from the trained U-Net model are found to be promising as shown in Fig.2. The segmented image from U-Net model gives us the quantitative information about the lesion region viz. area, perimeter, minor axis, major axis and colour information for measuring the ABCD parameter [15] of the dermatology such as Asymmetry, Border irregularity, Colour variation and Diameter for proper clinical diagnosis and treatment.
After image segmentation and cropping, the cropped image data becomes more meaningful representation as the normal skin portion is completely removed. Due to this reason, the LeNet-5 classifier model showed performance improvement with segmented lesion images when compared with unsegmented skin images as shown in Table-2. Thus the proposed algorithm having U-Net image segmentation model along with LeNet-5 classifier showed better classification results. Our proposed algorithm yielded maximum training accuracy of 93.80% and testing accuracy of 82.42% with maximum F1 score of 0.8384 and minimum misclassification rate of 17.57% when compared with other CNN models viz. LeNet-5 and ResNet-50 trained on unsegmented input image dataset. It is observed that after image segmentation, the LeNet-5 CNN model with its simple architecture and limited weights able to achieve better classification performance in terms of overall accuracy [16]. However, deeper CNN architectures like ResNet-50 could not show any performance improvement after lesion segmentation. The self-learning AI framework integrates the capability of unsupervised machine learning techniques with deep learning.

Table 1. Confusion Matrix of ResNet-50, LeNet-5 & the Proposed algorithm

| *ResNet-50* | | Actual | |
|---|---|---|---|
| **Predicted** | | Benign | Malignant |
| | Benign | 240 | 7 |
| | Malignant | 120 | 293 |

| *LeNet-5* | | Actual | |
|---|---|---|---|
| **Predicted** | | Benign | Malignant |
| | Benign | 261 | 22 |
| | Malignant | 99 | 278 |

| *Proposed Algorithm* | | Actual | |
|---|---|---|---|
| **Predicted** | | Benign | Malignant |
| | Benign | 301 | 57 |
| | Malignant | 59 | 243 |

Table 2. Performance Metrics of Different CNN Models

| Model | Training Accuracy (%) | Testing Accuracy (%) | F1 Score | Misclassification Rate (%) |
|---|---|---|---|---|
| Classification without segmentation | | | | |
| ResNet-50 | 81.00 | 80.75 | 0.7908 | 19.24 |
| LeNet-5 | 88.40 | 81.66 | 0.8118 | 18.33 |
| Classification with segmentation | | | | |
| *Proposed Algorithm* | 93.80 | 82.42 | 0.8384 | 17.57 |

## 5. CONCLUSIONS

In this paper, we proposed a self-learning AI framework for skin lesion region segmentation and classification in dermoscopy images. The U-Net segmentation model along with annotation scheme provides a novel self-learning solution for lesion region segmentation on the image dataset, which do not have ground truth information. To generate initial labeling data for training the U-Net model, the annotation scheme uses unsupervised clustering technique. This technique enormously reduces the expertise and time constraints involved in manual labeling. The proposed algorithm takes the skin image as input and segments the lesion region with trained U-Net model. The LeNet-5 classifier model was trained on these segmented images to classify them as benign or malignant. It was noted that after U-Net segmentation, the performance of the LeNet-5 classifier has been improved. Thus the overall classification accuracy of the proposed AI framework is much better than the other two CNN classifier models which are directly trained on the input dataset. The performance of the self-learning framework is effective in implementing end to end image processing solution for unlabelled skin lesion image dataset. This framework can also be extended to other medical image datasets like MRI images of brain tumour.

## ACKNOWLEDGEMENTS

We are sincerely thankful to the Director, CSIR-Electronics Engineering Research Institute, Pilani and Scientist-In Charge, CEERI Chennai Centre for providing the opportunity to carry out this work.